\documentclass[runningheads]{llncs}
\usepackage{graphicx}
\usepackage{amsmath}
\usepackage[ruled]{algorithm2e}
 \usepackage{amsfonts}
\usepackage{booktabs}
\usepackage{adjustbox}
\usepackage{multirow}
\begin{document}

\title{Generalizing Nucleus Recognition Model in Multi-source Ki67 Immunohistochemistry Stained Images via Domain-specific Pruning}
%
\titlerunning{Generalizing Nucleus Recognition Model in Multi-source Images via Pruning}
\author{Jiatong Cai \inst{1, 2} \and
Chenglu Zhu \inst{1, 2} \and
Can Cui \inst{1, 2} \and
Honglin Li \inst{1, 2} \and
Tong Wu \inst{1, 2} \and
Shichuan Zhang \inst{1, 2, 3} \and
Lin Yang \inst{1, 2}}
\authorrunning{J. Cai et al.}
\institute{Artificial Intelligence and Biomedical Image Analysis Lab, School of Engineering, Westlake University \and
Institute of Advanced Technology, Westlake Institute for Advanced Study \and
College of Computer Science \& Technology, Zhejiang University\\
\email{yanglin@westlake.edu.cn}}
\maketitle
\begin{abstract}
  Ki67 is a significant biomarker in the diagnosis and prognosis of cancer, whose index can be evaluated by quantifying its expression in Ki67 immunohistochemistry (IHC) stained images. However, quantitative analysis on multi-source Ki67 images is yet a challenging task in practice due to cross-domain distribution differences, which result from imaging variation, staining styles and lesion types. Many recent studies have made some efforts on domain generalization (DG), whereas there are still some noteworthy limitations. Specifically in the case of Ki67 images, learning invariant representation is at the mercy of the insufficient number of domains and the cell categories mismatching in different domains. In this paper, we propose a novel method to improve DG by searching the domain-agnostic subnetwork in a domain merging scenario. Partial model parameters are iteratively pruned according to the domain gap, which is caused by the data converting from a single domain into merged domains during training. In addition, the model is optimized by fine-tuning on merged domains to eliminate the interference of class mismatching among various domains. Furthermore, an appropriate implementation is attained by applying the pruning method to different parts of the framework. Compared with known DG methods, our method yields excellent performance in multiclass nucleus recognition of Ki67 IHC images, especially in the lost category cases. Moreover, our competitive results are also evaluated on the public dataset over the state-of-the-art DG methods.

\end{abstract}

\section{Introduction}
In the diagnosis and prognosis of cancer, Ki67 is a significant biomarker \cite{reis2020ki67,yang2018ki67,yerushalmi2010ki67} that can evaluate tumor cell proliferation and growth by quantifying its expression in Ki67 immunohistochemistry stained images\cite{kloppel2018ki67,volynskaya2019ki67}.
With the increase of data volume, deep learning methods have provided strong advantages in nucleus recognition \cite{xie2018efficient,xing2019pixel}. However, the lack of standardized multicenter data could bring heterogeneous Ki67 interpretations from different stainings, scanners and environments \cite{focke2017interlaboratory,rimm2019international}. Applying methods trained on a single domain can result in a biased prediction.
It is commonly required in practice that the robust model trained on multi-source data should have a reliable prediction on unseen domains. Empirical Risk Minimization (ERM) \cite{vapnik1999overview} is a convenient solution, however, the gap still exists due to the complexity of the merged domains.

Methods of cross-domain training can be simply divided into domain adaptation (DA) and domain generalization (DG). Specifically, DA aims to solve the covariate shift between the source domain and target domain.
A popular representation learning approach named domain-adversarial neural networks (DANN) was proposed by setting a domain discriminator to confuse domain features in an indiscriminate mechanism \cite{ganin2016domain}. In histopathological image analysis, DANN was applied to organ-level, slide-level \cite{lafarge2019learning} and patient-level domains \cite{hashimoto2020multi}.
Similarly, adversarial discriminative domain adaptation (ADDA) \cite{tzeng2017adversarial} is achieved by replacing the weight-shared backbone with exclusive domain-specific encoders, which was introduced into the quality assessment task of fundus images \cite{shen2020domain}. Besides, some studies implemented unsupervised DA by utilizing generative adversarial networks in liver segmentation \cite{yang2019unsupervised}, microscopy image quantification \cite{xing2019adversarial} and diabetic retinopathy detection \cite{yang2020residual}. Yet, since the model is barely trained on the source domain and target domain, the performance is flawed on unseen domains.

To achieve a desirable generalization performance, data augmentation is an intuitive strategy. For instance, adversarial samples are generated by adaptive data augmentation and deep-stacked transformation in \cite{volpi2018generalizing} and \cite{zhang2020generalizing}, respectively. Nevertheless, model convergence could be influenced by the generated unrealistic data. Instead, some researchers studied DG with learning domain-agnostic representations.
Albuquerque et al. improve generalization by minimizing discrepancies of the pairwise domain \cite{albuquerque2019generalizing}. Sikaroudi et al. utilize meta-learning to train a magnification-agnostic histopathology image classification model \cite{sikaroudi2021magnification}. Mahajan et al. give a causal interpretation to DG and generalize based on invariant representations \cite{mahajan2020domain}. The works above achieved optimistic results on data from unseen domains whereas there are still some non-negligible limitations. First of all, DANN-based methods could not function well with limited data in the source domain. What’s more, some studies are not applicable when certain categories do not exist overall training domains.

Based on this, we present a novel DG approach for nucleus recognition in Ki67 immunohistochemistry stained images. The proposed method transforms DG into a domain-agnostic subnetwork searching problem. Invariant representations are learned by iteratively pruning domain-specific model parameters under a domain merging scenario. Moreover, the pruned model is fine-tuned on merged domains, which solves the class mismatch problem to some extent. Furthermore, the appropriate implementation is achieved by applying our pruning method in different modules of the nucleus recognition framework. Compared with the state-of-the-art (SOTA) methods in the Ki67 IHC image and public data set, our method achieves competitive performance in the unseen domains.

\section{Methods}
\subsection{Overview}
\subsubsection{Motivation.}
The lottery ticket hypothesis \cite{frankle2018lottery} claims only a subset of parameters account for model performance in an over-parameterized neural network. Without any degradation of performance, the subnetwork compression can be still achieved by iteratively pruning trivial parameters. Under some circumstances, the model may not generalize well to unseen domains because of overfitting on domain-specific representations. Consequently, DG could be considered as a domain-agnostic subnetwork searching problem, which keeps domain-agnostic parameters while pruning domain-specific parameters.

\subsubsection{Workflow.}
\begin{figure}
\includegraphics[width=\textwidth]{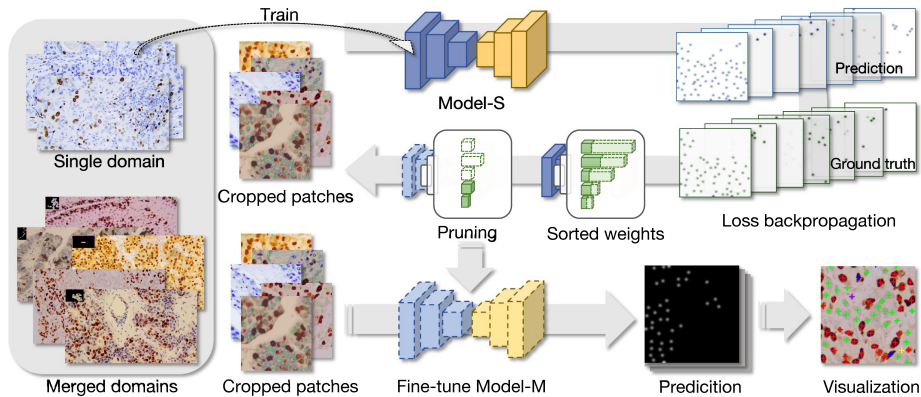}
\caption{A workflow of prune-based generalization. We use model-S and model-M to represent the model converged on a single domain and model pruned on merged domains.} \label{workflow}
\end{figure}

The proposed prune-based DG framework is illustrated in Fig.\ref{workflow}.
Initially, parameters ${W}$ are obtained by the convergent model (Model-S) trained on a single domain. When training in merged domains, model gradients are collected and sorted globally by their magnitude after the loss backpropagation. Subsequently the subset of ${W}$ will be achieved by pruning weights with the top ${p\%}$ gradient magnitude and resetting the remaining parameters without updating in ${W}$ by iterating n times. Finally, the pruned model (Model-M) will be fine-tuned in merged domains.

\subsection{Pruning-based Algorithm}

Our method aims to filter out a fraction of parameters that the remaining parameters ${W}'$ could preserve the accuracy on the training domains while performing well on the unseen domain(s). To achieve this goal, we construct a domain merging scenario.
That is, the model is firstly trained on a single source domain. After it is converged, we replace the training data with merged domains, which contain the source domain and domain(s) never seen before. For the latter, we denote it as invasion domain(s). Under this scenario, the target problem corresponds to the optimization as follows.
\begin{equation}\label{eq1}
  W^{*}=\mathop{\arg\min}\limits_{{W}'\subseteq {W}}\left | C(D_{s}|{W}')- C(D_{s}|W)\right |+C(D_{i}|{W}'),
\end{equation}
where $D_{s}$ and $D_{i}$ are samples from the source domain and invasion domains respectively, $W^{*}$ is the expected ${W'}$, and ${W}$ denotes the parameters trained on $D_{s}$.
The cost $C(D|W)$ was computed on samples of $D$ under parameters ${W}$. Generally, $C(D_{s}|{W}')$ of pruned model is larger than $C(D_{s}|W)$. The Eq. \ref{eq1} can be simplified as follows:
\begin{equation}\label{eq2}
  W^{*}=\mathop{\arg\min}\limits_{{W}'\subseteq {W}}C(D_{m}|{W}')- C(D_{s}|W) ,\ \text{where} \ D_{m} = \{D_{s}, D_{i}\} \, ,
\end{equation}
${D_{m}}$ represents a batch of merged samples that consist of ${D_{s}}$ and ${D_{i}}$. Subsequently, the second term of Eq. \ref{eq2} has no contribution to the selection of ${W}$ thus can be removed, and the optimization of ${W'}$ can be defined as Eq. \ref{eq3},
\begin{equation}\label{eq3}
  W^{*}=\mathop{\arg\min}\limits_{{W}'\subseteq {W}}C(D_{m}|{W}').
\end{equation}
The above subnetwork searching problem can be solved by exploring the contribution of each parameter ${w}$ to ${C(D_{m}|W)}$. More concretely, while trained on $D_{m}$, model parameters ${W}$ are updated through loss backpropagation according to the gradient descent. In this procedure, gradients are calculated as derivatives of the cost with respect to ${W}$ as $g={\partial C(D_{m}|W)}/{\partial W}$. Correspondingly, the gradient magnitude ${\left |{g} \right |}$ of the individual parameter ${w\in W}$ can be regarded as its contribution to ${C(D_{m}|W)}$. Intuitively, model convergence depends on the updating of the parameters. Parameters requiring a large variation to fit on merged domains can be considered highly domain-related. The expected remaining parameters $W^{*}$ can be approximated by preserving parameters with a relatively small ${\left |g\right |}$ as Eq. \ref{eq4},
\begin{equation}\label{eq4}
   W_r=\left\{\begin{matrix}
   w_{i} ,& \text{ if } g_{i}\leq c_{p} \\
   0 ,& \text{otherwise}
 \end{matrix}\right.,
 \end{equation}
where $c_{p}$ is a scalar related to remaining percentage $p$ of parameters.

Generally, a rather small pruning rate is set for preventing performance degradation by using a large pruning rate. Moreover,  the same setting as \cite{frankle2018lottery}  is adopted through iterative training and pruning followed by resetting the remaining parameters over n steps with each step prunes ${p^{\frac{1}{n}}\%}$ of the surviving parameters. A few iterations of model pruning is usually enough, therefore a small amount of merged data is enough in the pruning procedure. Although the pruned model suffers degraded performance, it is provided with the potentiality of learning domain-invariant representations. Finally, further fine-tuning is applied to surviving parameters on merged domains for recovering accuracy.

\section{Experiments}
\subsection{Nucleus Recognition}
\subsubsection{Ki67 Dataset.}
All Ki67 IHC images were collected under magnification $40\times$ and resolution $1920\times1080$ from 16 domains through data anonymization. Nuclei of each slice were annotated by certified pathologists in seven categories: negative fibroblasts, negative lymphocyte, negative tumor cell, positive fibroblasts, positive lymphocyte, positive tumor cell and other cells. The data set was built under the mismatched categories in different domains as shown in Table.\ref{ki67data}.
\begin{table}\centering
\caption{Properties of different domains in Ki67 data set}\label{ki67data}{
\begin{tabular}{c|c|c|c}
\hline
\multirow{2}{*}{Properties}  & \multicolumn{2}{c|}{Merged training domains} & \multirow{2}{*}{Testing unseen domains} \\
     & \multicolumn{2}{c|}{\ \ D1\ \ \ \ \ \ \ \ \ \ \ \ \ D2\ \ \ \ } &  \\ \hline
No. of images & 95  & 51  &  41 \\\hline
No. of categories  & 7 &  5  & 4,5,7  \\\hline
Neuclei count & 35,948 & 35,948 & 35,948 \\\hline
Radius of cells & 4$\sim$6 pixel & 7$\sim$12 pixel& 5$\sim$14 pixel \\\hline
Type of tumor &  breast & breast & breast, cervix, pancreas, urethra  \\ \hline
Hue & partial blue & partial orange & partial violet, brown red, yellow green \\\hline
\end{tabular}
}
\end{table}

\subsubsection{Settings and Implementations.}
Nucleus recognition was formulated to a structured regression problem by implementing a weakly supervised segmentation \cite{xing2019pixel}. The ground truth mask is a set of continuous-valued proximity maps ranging from 0 to 1, generated by applying a 2D Gaussian filter on the annotations. Image patches are randomly cropped with resolution $512\times512$ and fed into the variant of Albunet (U-Net \cite{ronneberger2015u} with ResNet34 \cite{he2016deep} pre-trained on ImageNet as the encoder). The model was trained by Adam optimizer under the sum of cross-entropy loss and IOU loss with adjusted hyper-parameters: batch size 8, iterations 500, initial learning rate $5\times10^{-4}$ with gradually decreasing after every 100 epochs. Our base model was pre-trained on D1 and pruned on the merged domain (D1 and D2) with pruning ratio $10^{-4}$, prune-reset iterations 4. Besides, an ablation experiment was arranged to observe the effect of pruning, which intervened the encoder, the decoder and all modules under the same pruning rate. In addition, data augmentation was also adopted including random rotation, crop, flip, mirroring and color jitter. The optimal model could be selected if the accuracy no longer improved within 30 epochs. The final prediction was achieved by non-maximum suppression in the probability map.
\begin{table}\centering
\caption{Nucleus Recognition DG results on unseen \& merge domains of Ki67}\label{ki67results}
\resizebox{\textwidth}{!}{
\begin{tabular}{c|c|c c c c|c c c}
\hline
\multirow{2}{*}{\bfseries Algorithm} &\multirow{2}{*}{Test (domains)} & \multicolumn{4}{c}{Detection} & \multicolumn{3}{|c}{Classificaion} \\

&  & \textit{P} & \textit{R} & \textit{F1} & \textit{${\mu \pm \sigma}$} & \textit{P} & \textit{R} & \textit{F1} \\
\hline
\multirow{2}{*}{ERM} &
{merge} & { $\mathbf{86.86}$ } & { $\mathbf{85.70}$ }  & { $\mathbf{86.28}$ }  & { ${5.59 \pm 3.17}$ } & { $\mathbf{86.68}$ } & { 82.78 }& { 83.97 }\\
& {unseen} &  { 89.56 } & { 82.92 }  & { 86.11 }  & { ${5.63 \pm 3.02}$ } & { 90.45 } & { 87.65 }& { 88.35 }\\
\hline
\multirow{2}{*}{ERM-F}
& {merge} & { 85.22 } & { 83.53 }  & { 84.37 }  & { ${5.70 \pm 3.15}$ } & { 84.43 } & { 84.10 }& { 84.19 }\\
&  {unseen} & { 88.66 } & { 81.75 }  & { 85.06 }  & { ${5.59 \pm 2.99}$ } & { 90.99 } & { 88.57 }& { 89.38 }\\
\hline
\multirow{2}{*}{DANN\cite{ganin2016domain}}
& {merge} & { 75.43 } & { 80.90 }  & { 78.07 }  & {${5.89 \pm 3.36}$} & { 77.63 } & { 72.16 }& { 74.09 }\\
 &  {unseen} & { 81.73 } & { 78.94 }  & { 80.31 }  & {${5.76 \pm 3.19}$} & { 86.84 } & { 68.31 }& { 74.77 }\\
\hline
\multirow{2}{*}{Adv-Aug\cite{volpi2018generalizing}}
& {merge} & {80.53} & { 60.65 }  & { 69.19 }  & {${6.39 \pm 3.47}$} & { 69.45 } & { 76.40 }& { 72.45 }\\
& {unseen} &  {$\mathbf{90.88}$} & { 61.92 }  & { 73.66 }  & {${5.64 \pm 3.07}$} & {88.17} & { $\mathbf{90.69}$ }  & { 89.14 }\\
\hline
\multirow{2}{*}{Ours-Encoder}
& {merge} &  { 85.30 }  & { 85.76 }  & { 85.53 }  & {${5.57 \pm 3.17}$} & { 84.93 } & { 84.24 }& { 84.39 }\\
& {unseen} &  { 89.26 }  & { 83.11 }  & { 86.08 }  & {${5.51 \pm 3.01}$} & { $\mathbf{91.81}$ } & { 89.03 }& { 89.72 }\\
\hline
\multirow{2}{*}{Ours-Decoder}
& {merge} &  { 84.86 }  & { 84.96 }  & { 84.91 }  & {${\mathbf{5.53 \pm 3.17}}$} & { 84.85 } & { 82.81 }& { 83.52 }\\
& {unseen} &  { 87.99 }  & { 83.01 }  & { 85.43 }  & {${\mathbf{5.45 \pm 3.02}}$} & { 91.37 } & { 87.82 }& { 88.89 }\\
\hline
\multirow{2}{*}{Ours-All}
& {merge} &  { 85.95 }  & { 85.19 }  & { 85.57 }  & { ${5.57 \pm 3.15}$ } & { 86.43 } & { $\mathbf{85.25}$ }& { $\mathbf{85.70 }$}\\
& {unseen} &  { 88.83 }  & { $\mathbf{84.67}$ }  & {$\mathbf{86.46}$ }  & { ${5.54 \pm 3.02}$ } & { 91.04 } & { 89.42 }& { $\mathbf{89.91}$ }\\
\hline
\end{tabular}}
\end{table}

\subsubsection{Evaluation.}
The results were evaluated on precision, recall and F1-score under the two tasks as follows. The final score is averaged over 41 cases.
\begin{itemize}
  \item Detection. Hungarian matching algorithm \cite{kuhn1955hungarian} was applied to the predictions and annotations under the valid area with a radius of 16 pixels to calculate true positive (TP), false positive (FP) and false negative (FN). The average distance $\mu$ and standard deviation $\sigma$ of paired results were computed as an additional indicator.
  \item Classification. The pair-wise results in detection were further assessed according to whether their categories are matched using metrics in \cite{sirinukunwattana2016locality}.
\end{itemize}
The competitive methods and results were shown in Table \ref{ki67results}. We take ERM-F(trained and fine-tuned the same as our method but without pruning) as a reference to guarantee the improvement is not gained from fine-tuning. Moreover, DANN\cite{ganin2016domain} was implemented by embedding a domain discriminator following the model encoder. Adv-Aug\cite{volpi2018generalizing} was adjusted with the adversarial learning rate ranging from 0 to $10^{5}$, step 10.
\subsubsection{Results and Discussions.}
Nucleus Recognition results are shown in Table \ref{ki67results}. ‘Merge’ and ‘unseen’ represent domains participating in training and never involved in training, respectively. ‘Ours-Encoder’ and ‘Ours-Decoder’ denote applying the pruning method only on the ‘encoder’ and ‘decoder’ modules. Our method exhibits the best comprehensive performance on merged domains and unseen domains. Considering the performance on unseen domains, metrics of recall and F1-score show competitive results of our method on both nucleus detection and classification. Particularly, the proposed method improve the detection F1-score by 0.35${\%}$ for ERM and 1.40${\%}$ for ERM-F and for nuclues classification our method outperforms ERM and ERM-F in terms of F1-score by 1.56${\%}$ and 0.53${\%}$, respectively. On merge domains, the proposed method slightly under-performs the ERM algorithm in nucleus detection. Such performance degradation is reasonable. Because ERM, with limited generalization, tends to over-fit on domain-specific features which may help with the accuracy of merge domains. Even so, our methods surpass ERM and ERM-F by a non-trivial margin of 1.73${\%}$ and 1.51${\%}$(F1-score), respectively on merge domains in classification.

It is worth noting that DANN shows rather worse metrics than ERM in our experiment, which indicates its vulnerability in coping with the class mismatching problem both on merged domains and the generalization on unseen domains. The expected effect of Adv-Aug is marginal on unseen domains because of the complexity of histopathological images. It is observed that model training interferes with the generated unrealistic images. The above two methods utilize adversarial mechanisms for the expansion of available domains, whose limitation appears when only a small number of domains is available for training. On the contrary, our pruning method attempts to learn the domain-agnostic representations by preventing the model's over-fitting on training domains.

\begin{figure}[htbp]\centering
\includegraphics[width=\textwidth]{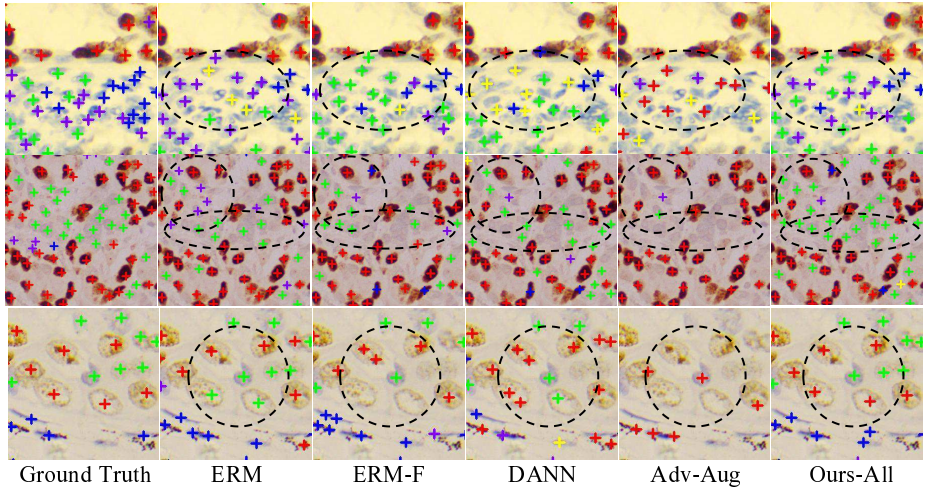}
\caption{Comparison of nuclei detection and classification with different methods} \label{visualization}
\end{figure}
Fig. \ref{visualization} shows representative Ki67 recognition results of the above methods in three distinct domains for comparison. Blue, yellow, green, red and violet represent negative fibroblasts, negative lymphocytes, negative tumor cells, positive tumor cells and other cells. The remaining two categories do not show up in these cases. Areas with a significant difference among methods are highlighted with black dash ellipses. Compared with other approaches, the proposed method produces relatively better results in classification (case 1, 3 in row 1 and 3) and detection (case 2 in row 2). To be specific, in case 1, our method shows competitive classification results given the mixture of negative tumors, negative fibroblasts and other cells. The improvements of detection are also noticeable in case 2 even though the color of negative tumor cells is too shallow to be figured out. Case 3 further shows the robustness of our method on hollow cells.

The ablation experiment explores the pruning availability in different modules. Generally, pruning on all modules achieves the best performance. This can be caused by the remaining percentage of parameters within each layer. It was observed that the proportion of remaining parameters has a wide variance in different network depths although pruned with a tiny rate. The Decoder-only approach prunes the final layer to a large sparsity (70${\% \sim 80\%}$) and Encoder-only reserves 30${\% \sim 50\%}$ parameters for shallow layers. By contrast, pruning on all modules preserves 40${\% \sim 50\%}$ parameters for the final layer and 40${\% \sim 60\%}$ parameters for the shallow layers. All evaluations illustrate the importance of allocating parameters to prune in a balanced manner.

\subsection{PACS Classificaion}
\begin{table}\centering
\caption{Domain generalization on PACS, Model selection: oracle}\label{tab2}
\begin{tabular}{c|c|c|c|c|c}
\hline
{Algorithm} &  {Art} & {Cartoon} & {Photo} & {Sketch} & {Average}\\
\hline
ERM \cite{vapnik1999overview}&  { $\mathbf{87.42}$ ${\pm}$ 0.4 } & { 82.56 ${\pm}$ 0.6}  & { 96.73 ${\pm}$ 0.4}  & { 80.02 ${\pm}$ 0.6} & { 86.68 }\\
IRM \cite{arjovsky2019invariant} &  { 86.37 ${\pm}$ 0.8} & { 80.12 ${\pm}$ 1.2}  & { 97.24 ${\pm}$ 0.4}  & { 76.38 ${\pm}$ 1.1} & { 85.02 }\\
DRO \cite{sagawa2019distributionally}&  { 87.53 ${\pm}$ 0.7} & { 81.74 ${\pm}$ 0.7}  & { $\mathbf{97.61}$ ${\pm}$ 0.2} & { 79.80 ${\pm}$ 0.8} & { 86.67 }\\
MLDG \cite{li2018learning}& { 87.17 ${\pm}$ 0.8} & { 81.35 ${\pm}$ 1.4}  & { 97.36 ${\pm}$ 0.4} & { 80.94 ${\pm}$ 1.0} & { 86.71 }\\
DANN \cite{ganin2016domain}& { 86.52 ${\pm}$ 1.2} & { 80.25 ${\pm}$ 1.1}  & { 97.58 ${\pm}$ 0.2}  & { 77.26 ${\pm}$ 1.2} & { 85.40 }\\
CORAL \cite{sun2016deep}& { 87.21 ${\pm}$ 0.6} & { 82.16 ${\pm}$ 0.4}  & { 97.52 ${\pm}$ 0.1}  & { 80.01 ${\pm}$ 0.3 } & { 86.77 }\\
$\mathbf{Ours}$ &  { 85.33 ${\pm}$ 0.5} & { $\mathbf{84.83}$ ${\pm}$ 0.6} & { 96.11 ${\pm}$ 0.4}  & { $\mathbf{83.95}$ ${\pm}$ 0.2} & { $\mathbf{87.56}$}\\
\hline
\end{tabular}
\end{table}
PACS (photo, art painting, cartoon, sketches) classification benchmark was also introduced to evaluate our method with ResNet50 backbone on test domains validation set (oracle) \cite{gulrajani2020search}. The results are shown in Table. \ref{tab2}. is the maximum value over 3 experiments, and within each experiment, the source domain and the invasion domain are arranged alternately in a ratio of 1:2. It is observed that our method boosts the performance on the sketches domain, outperforming the SOTA methods by 3.01${\%}$. The proposed method also surpasses other approaches on the domain of cartoon by a margin of 2.27${\%}$. Note that skeletons in sketches and contours in cartoons are cross-domain structural information. Therefore, the above results could well demonstrate the effectiveness of our approach in learning domain-agnostic representations. The potential reason for our mediocre results on the photos might be the pruning method is applied on the model pre-trained on the photo domain (ImageNet). Consequently, the corresponding results are biased and have limited reference value. In addition, the factor that leads to our imperfect performance is probably the huge variance between the art painting domain and other domains. Even so, our method still achieves the best performance in the average accuracy.

\section{Conclusion}
In this paper, we propose a novel DG method based on subnetwork searching, which generalizes the model by discarding the domain-specific representations and reserving the invariant representations in a domain merging scenario. Our method attempts to solve practical limitations of DG including the insufficient number of domains and the class mismatching problem cross domains. Competitive results demonstrate our method's superiority and robustness on multiclass nucleus recognition in Ki67 IHC images. In addition, the proposed method achieves good performance by effectively learning generic representations the on PACS classification benchmark.

\bibliographystyle{splncs04}
\bibliography{paper1802.bib}
\end{document}